\documentclass[letterpaper, 10 pt, conference]{ieeeconf}

\IEEEoverridecommandlockouts

\usepackage{amsmath}
\usepackage{amssymb}
\usepackage{graphicx}
\usepackage{amsmath}
\usepackage{float}
\usepackage{url}
\usepackage{soul, color, xcolor}
\usepackage{booktabs} 
\usepackage{multirow}
\usepackage{subfig}
\usepackage{hyperref}

\title{\LARGE \bf
SKATER: Synthesized Kinematics for Advanced Traversing Efficiency on a Humanoid Robot via  Inline-Skate Swizzles
}

\author{Junchi Gu, Feiyang Yuan, Weize Shi, Tianchen Huang, Haopeng Zhang, Xiaohu Zhang, \\Yu Wang, Wei Gao and Shiwu Zhang%
\thanks{This work was supported in part by the National Natural Science Foundation of China under Grants U22B2040, in part by the Major Project of Anhui Province's Science and Technology Innovation Breakthrough Plan (202423h08050003), in part by the Key Science \& Technology Project of Anhui Province (202523j08050001), and in part by the Fundamental Research Funds for the Central Universities with grant No. YD2090002019. (Corresponding authors: Wei Gao; Yu Wang.)}%
\thanks{The authors are with the Institute of Humanoid Robots, Department of Precision Machinery and Precision Instrumentation, University of Science and Technology of China, Hefei, Anhui 230026, China. {\tt\footnotesize weigao@ustc.edu.cn; wangyuustc@ustc.edu.cn}}%
}

\begin{document}

\maketitle
\thispagestyle{empty}
\pagestyle{empty}

\begin{abstract}

Humanoid walking and running have advanced rapidly, but their intermittent ground contacts still produce large transient loads at touchdown and cause undesired energy loss. This paper presents a humanoid robot, SKATER, whose feet are equipped with passive inline wheels, and investigates the swizzling gait as a continuous rolling-contact locomotion primitive for advanced efficiency. Because the wheels are neither actuated nor steerable, propulsion must be generated through whole-body motion to take advantage of the anisotropic and nonholonomic wheel-ground constraints. We formulate swizzling as a reference-free deep reinforcement learning problem, where neither motion-capture demonstrations nor predefined reference motions are necessary. Instead, reward design regarding ZMP criterion, inter-foot geometry, bilateral symmetry and wheel axial-slip constraints can guide the learning process without explicit gait-phase scheduling. Ablation study results quantify the contribution of these constraints, and the learned policy is successfully deployed on the physical robot. Against a walking baseline on the same platform with rigid flat feet, the swizzling gait reduces Impact Intensity based on ground reaction forces by $50.25\%$ in simulation, and mechanical Cost of Transport by $63.34\%$ in real-world experiments. Additional hardware tests yield $10/10$ successful $10$-m swizzling trials on surfaces including tile, rubber and gravel, as well as dynamic maneuver in obstacle avoidance. In high-speed trials, the robot can reach an average speed of $2.99 \pm 0.01$~m/s. The video is available here:\href{https://youtu.be/c9pRznL5lpI}{Video Link}

\end{abstract}

\section{Introduction}

Humanoid locomotion has progressed rapidly from walking~\cite{gu2024humanoid} and running~\cite{li2025reinforcement} to jumping~\cite{he2025asap} and whole-body skills~\cite{zhang2025track,han2025kungfubot2}. Nevertheless, walking and running still involve repeated foot-ground impacts that incur large ground reaction forces (GRFs) and energy losses, thus increase joint loads and acoustic noises. Inline skating replaces discrete touchdown with continuous rolling contact, which has been reported by human biomechanics studies as having lower impact loads than running while retaining efficient forward motion~\cite{mahar1997impact}. This motivates inline skating with passive wheels as a complementary mobility mode for humanoid robots on extended flat terrain.

\begin{figure}[t]
    \centering
    \includegraphics[width=1.0\columnwidth]{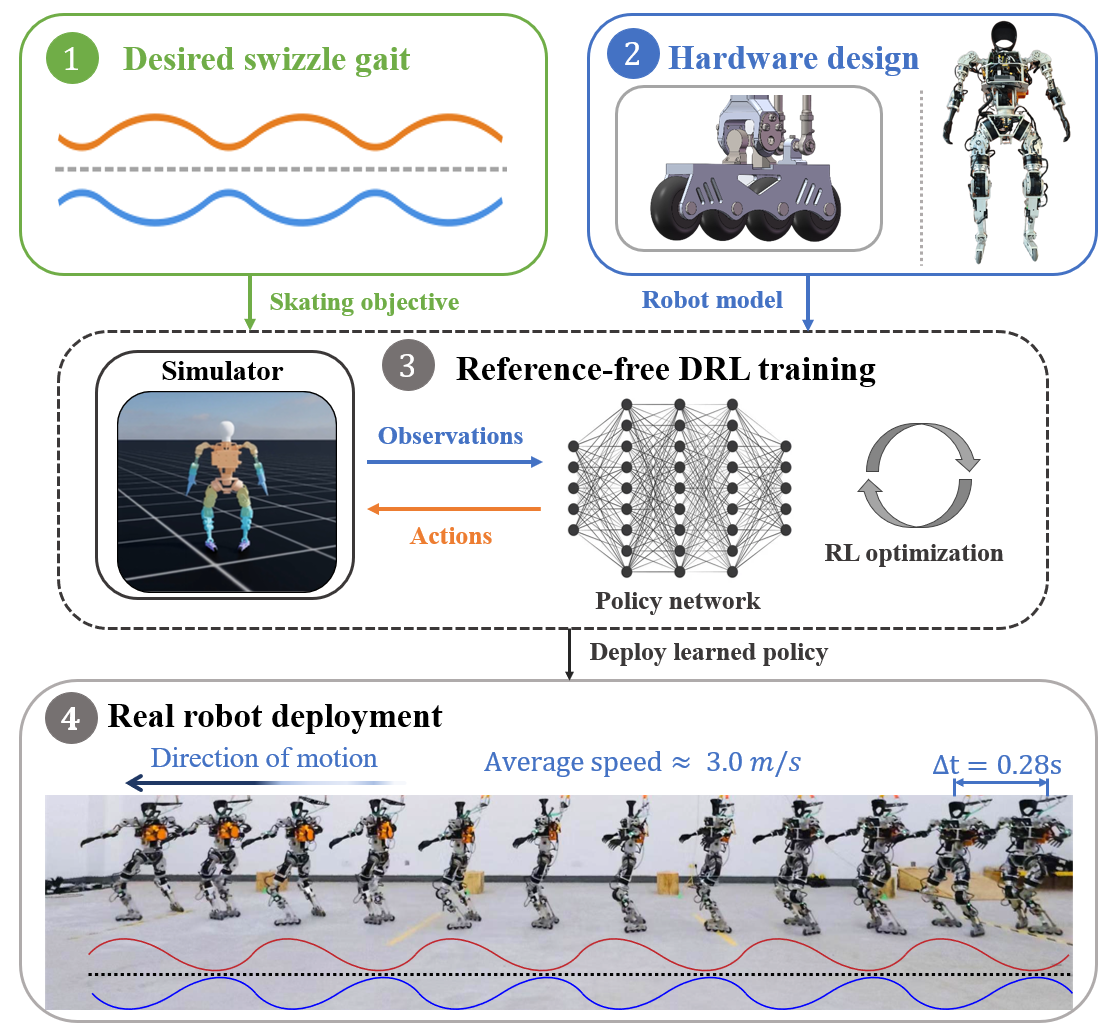}
    \caption{Overview of the SKATER platform and its learned swizzling gait.}
    \label{fig:System diagram}
\end{figure}

Realizing this capability is challenging because passive inline wheels are neither actuated nor steerable. Propulsion and maneuvering must instead arise through whole-body motion under the anisotropic and nonholonomic wheel-ground contacts, while the controller maintains balance and suppresses lateral slip. Swizzling provides a suitable entry-point primitive: periodic bilateral abduction and adduction of the feet generate thrust while maintaining continuous support. The controller must therefore coordinate ground contact geometry, wheel heading, support regulation and trunk stabilization without a predefined contact schedule.

Existing studies only partially address this problem. Early bipedal skating systems rely primarily on planned trajectories or model-based control~\cite{5708312,hashimoto2008swizzle,iverach2014gait,8450031}, whereas quadrupedal skating exploits different morphologies and balance mechanisms~\cite{bjelonic2018skating,chen2024unlocking}. On the other hand, learning-based humanoid control has achieved robust locomotion performance~\cite{tang2025deep,gu2024humanoid}, but often requires foothold, contact-schedule or gait-phase priors~\cite{lee2024integrating,gu2024humanoid}. Although hierarchical learning can reduce dependence on full-body motion references~\cite{peng2017deeploco}, learning continuous bilateral-support swizzling without reference motion data on a humanoid robot with passive inline wheels remains sparsely studied.

To address this gap, we present SKATER (Synthesized Kinematics for Advanced Traversing Efficiency on a humanoid Robot), a $33$-degree-of-freedom (DoF) humanoid robot with passive inline-wheel feet, as shown in Fig.~\ref{fig:System diagram}, and a reference-free and constraint-guided deep reinforcement learning (DRL) framework. Here, \emph{reference-free} denotes training without motion-capture demonstrations or predefined reference motions. The policy instead uses only proprioceptive history, skating-specific physical constraints and domain randomization, without any explicit gait-phase scheduling.

The main contributions of this paper include:

\begin{itemize}
\item We develop SKATER, a $33$-DoF humanoid platform with four passive inline wheels at each sole for studying continuous rolling-contact locomotion.

\item We formulate swizzling as a reference-free DRL problem that requires no reference motion data. Reward design regarding zero moment point (ZMP) criterion, inter-foot geometry, bilateral symmetry and wheel axial-slip suppression can guide learning without explicit gait-phase scheduling.

\item We validate the learned swizzling policy in simulation and on hardware. Relative to a separately trained walking baseline on the same robot but with rigid flat feet, swizzling reduces GRF-based Impact Intensity by $50.25\%$ in simulation and mechanical CoT by $63.34\%$ in real world. In high speed trials, swizzling can reach an average speed of $2.99 \pm 0.01$~m/s repeatedly.

\end{itemize}

\section{Related Work}
\label{Related Work}

\subsection{Skating Locomotion in Legged Robots}
\label{Roller Skating Locomotion in Legged Robots}

Passive-wheel skating in legged robots has primarily been addressed through trajectory planning and model-based control. Itabashi \emph{et al.} generate bipedal swizzling trajectories using fifth-order Bézier curves and a variable-curvature skate mechanism~\cite{5708312}. Hashimoto \emph{et al.} model anisotropic rolling and lateral friction, then modulate a foot reference trajectory from GRF feedback to limit internal forces~\cite{hashimoto2008swizzle}. Iverach-Brereton \emph{et al.} incorporate forward GRFs into model-based gait design for skating on solid and even ice ground~\cite{iverach2014gait}. These physically interpretable approaches establish the feasibility of bipedal skating and explicitly regulate contact behaviors. Across these studies, the propulsion cycle is prescribed before execution, while feedback mainly stabilizes or schedules the planned motion. This separation provides predictable behavior, but limits adaptation when surface friction, commanded velocity or robot dynamics depart from the design conditions. Extending these approaches to higher-dimensional humanoids and broader command ranges therefore requires substantial trajectory and model redesign.

Work on quadrupedal robots has advanced passive-wheel contact modeling and whole-body control. Bjelonic \emph{et al.} replace the conventional friction cone with a friction triangle and use hierarchical force control on ANYmal, reducing CoT by up to $80\%$ relative to walking~\cite{bjelonic2018skating}. Chen \emph{et al.} combine a geometric wheel-ground contact model with a Levenberg--Marquardt controller on a $4$-DoF wheel-legged quadruped, enabling diverse skating gaits and small-radius turns~\cite{chen2024unlocking}. The friction-triangle formulation captures direction-dependent traction explicitly, whereas the geometric wheel model improves contact-point accuracy in skating kinematics. Both developments are important for passive wheels, but their controllers remain tied to morphology-specific force distributions and planned motions. Moreover, quadrupedal skating benefits from a larger support region, redundant contacts and different load distribution. Its planning and stability mechanisms therefore do not transfer directly to a humanoid, leaving learning swizzling on a full-size humanoid without reference motion data unexplored.

\subsection{Locomotion Control of Humanoid Robots}
\label{Gait Generation and Learning-Based Control for Humanoid Robots}

The ZMP criterion~\cite{vukobratovic2004zero} and Model Predictive Control~\cite{katayama2023model} provide interpretable foundations for humanoid locomotion, but typically require accurate dynamics, planned CoM or foothold trajectories, and predefined contact sequences. Model-based planning makes stability constraints transparent and allows contact forces to be optimized directly. For a wheel-footed humanoid, however, it requires jointly modeling high-dimensional whole-body dynamics, and changing wheel headings, anisotropic frictions and multiple rolling contacts throughout the propulsion cycle. These assumptions are effective when stance and swing phases are known, but passive inline wheels remain in continuous contact and admit strongly direction-dependent motion. 

Learning-based methods improve adaptability while often preserving similar structure. Lee \emph{et al.} use a Linear Inverted Pendulum model to generate desired footholds for an RL tracking policy~\cite{lee2024integrating}. DeepLoco learns robust bipedal skills with limited prior structure, yet still relies on step targets and style objectives~\cite{peng2017deeploco}. Humanoid-Gym enables large-scale training and zero-shot sim-to-real transfer, while explicitly encoding gait phase and expected foot-contact patterns~\cite{gu2024humanoid}. Such priors improve learning efficiency and gait regularity, but constrain the learned motion to selected footfall structures and do not directly represent continuous and nonholonomic rolling contact. This distinction matters because swizzling speed emerges from coupled lateral foot motion and wheel orientation rather than choosing when and where to lift and place a foot.
Neither discrete foothold targets nor an alternating support schedule naturally describes the propulsion cycle. Therefore, this work adopts a reference-free formulation that learns without reference motion data, using support, geometric, symmetry and rolling constraints instead of explicit gait-phase scheduling.


\section{Methodology}
\label{METHODOLOGY}

\subsection{Hardware Design}
\label{Hardware Design}

\begin{figure}[t]
    \centering
    \includegraphics[width=1.0\columnwidth]{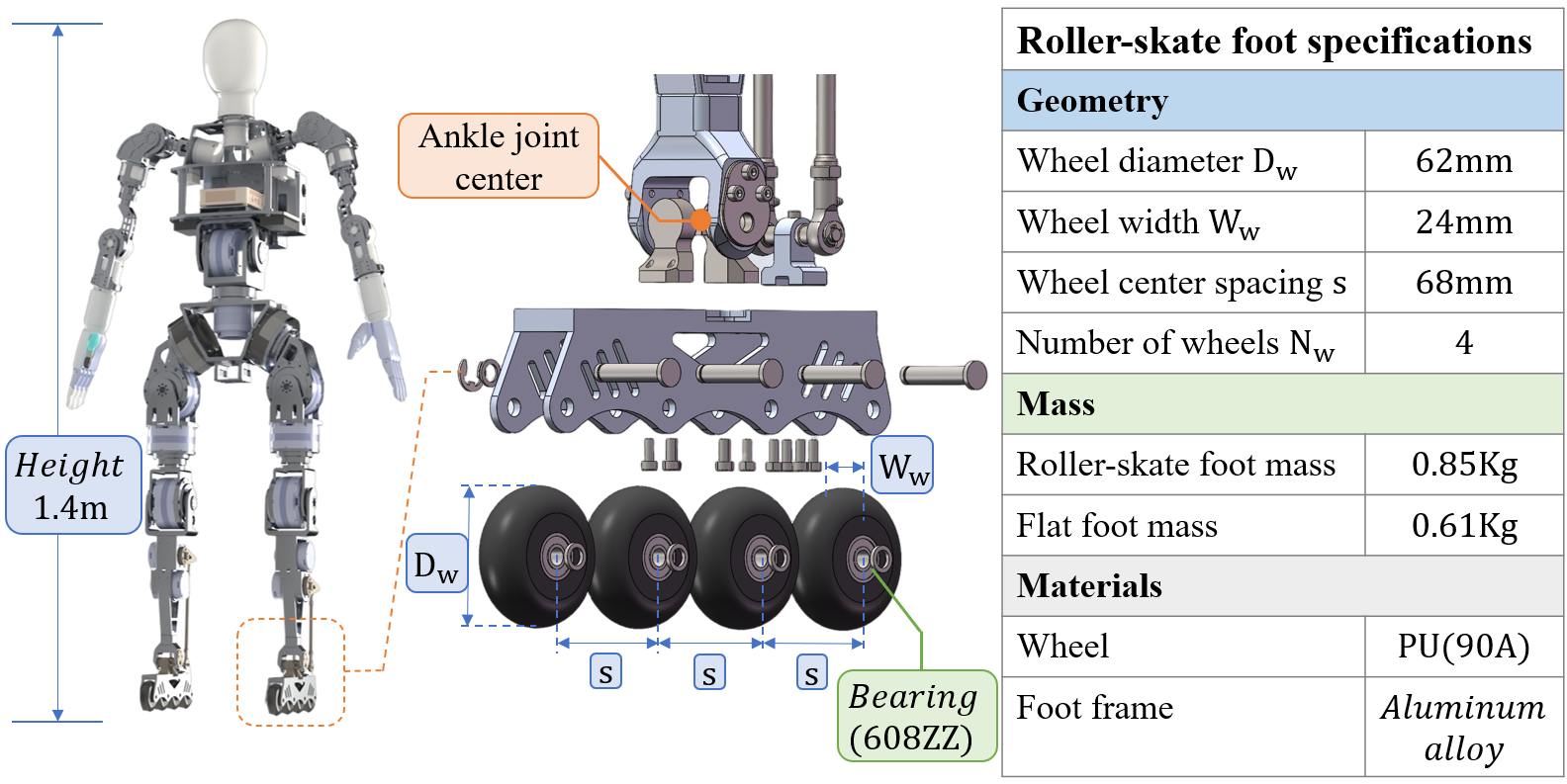}
    \caption{Hardware design of SKATER and its passive-wheel inline-skate foot module.}
    \label{fig:hardware introduction}
\end{figure}

As mentioned, SKATER is a humanoid robot with $33$ DoFs in total. The $25$ actuated DoFs include $6$ per leg, $5$ per arm, $2$ in the waist and $1$ in the neck, as shown in Fig.~\ref{fig:hardware introduction}. Each foot is equipped with $4$ inline wheels. The robot has a total mass of $38$ kg and a standing height of $140$ cm. 
Fig.~\ref{fig:hardware introduction} further shows the ankle-joint location and the exploded foot assembly. The inline-skate foot module uses polyurethane wheels with diameter $D_w=62$~mm, width $W_w=24$~mm and center spacing $s=68$~mm. Each inline-skate foot module has a mass of $0.85$~kg, while the rigid flat foot used for the walking baseline has a mass of $0.61$~kg. The wheels are mounted on passive 608ZZ bearings, and the foot frame is made of aluminum alloy.

Here, the inline skate shoes are not directly utilized due to two critical considerations:
1) Inline skate shoes exhibit incompatibility to nominal rigid flat foot, because their outer shells introduce a compliant layer to the foot sole and can cause relative sliding, which compromises the accuracy of force transmission and position control. 2) Roller skate shoes can interfere with the transmission mechanism of the robot's ankle joints, limiting its range of motion.
Another point worth noting is the wheel diameter.
Comparing to the wheel diameters between $72$-$80$ mm that are commonly used on standard inline skate shoes, the smaller wheel diameter of $62$ mm can lower the CoM of both feet and reduce the torque required around the ankle roll axis. This optimization is particularly critical during the lateral propulsion phase of inline skating, where the ankle joint must withstand significant eversion and inversion moments. 

\subsection{Reinforcement Learning of Swizzling}

\begin{figure*}[t]
    \centering
    \includegraphics[width=1.0\textwidth]{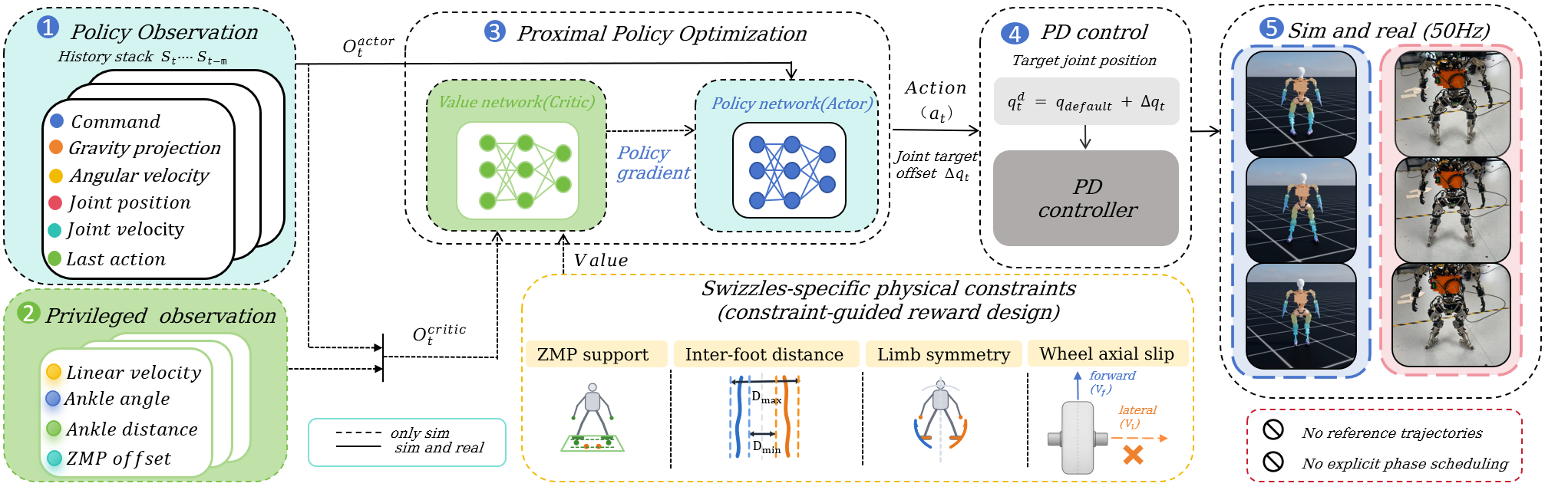}
    \caption{Deep reinforcement learning control framework for SKATER. The policy network processes proprioceptive sensor data to generate joint-level commands, enabling adaptive inline skate swizzles.}
    \label{fig:system}
\end{figure*}

\subsubsection{Problem Formulation}

We model the simulated swizzling task as a finite-horizon MDP~\cite{puterman2014markov} and optimize a stochastic policy $\pi_\theta(\boldsymbol{a}_t|\boldsymbol{o}_t)$ by maximizing the expected discounted return $J(\theta)=\mathbb{E}_{\pi_\theta}[\sum_{t=0}^{T-1}\gamma^t r_t]$, where $T$ is the episode horizon and $\gamma \in [0, 1]$ is the discount factor. The policy is trained with PPO~\cite{schulman2017proximal} using an asymmetric actor--critic in Isaac Lab~\cite{mittal2025isaac} with $4096$ parallel environments. 
Command ranges and domain-randomization magnitudes are expanded progressively to stabilize exploration.

\subsubsection{Observation Space}

As shown in Fig.~\ref{fig:system}, the Actor uses a five-step history of proprioceptive observations:
\begin{equation}
\boldsymbol{o}_t^{\text{actor}} = [\boldsymbol{s}_t, \boldsymbol{s}_{t-1}, \boldsymbol{s}_{t-2}, \boldsymbol{s}_{t-3}, \boldsymbol{s}_{t-4}]
\end{equation}
where the current state is defined as
\begin{equation}
\boldsymbol{s}_t = [\boldsymbol{q}_t, \dot{\boldsymbol{q}}_t, \boldsymbol{\omega}_t, \boldsymbol{g}, \boldsymbol{v}_t^{\text{cmd}}, \boldsymbol{a}_{t-1}]
\end{equation}
Here $\boldsymbol{q}_t,\dot{\boldsymbol{q}}_t \in \mathbb{R}^{n_j}$ are the actuated joint positions and velocities, with $n_j=25$ for SKATER, $\boldsymbol{\omega}_t \in \mathbb{R}^3$ is the base angular velocity, $\boldsymbol{g} \in \mathbb{R}^3$ is the gravity vector projected in the base frame, $\boldsymbol{v}_t^{\text{cmd}} \in \mathbb{R}^3$ is the commanded horizontal velocity and yaw rate, and $\boldsymbol{a}_{t-1} \in \mathbb{R}^{n_j}$ is the previous action. The eight passive wheel DoFs are excluded from both Actor and Critic observations because they are not actuated; Their rolling and slip effects are inferred from proprioceptive history rather than direct wheel-state inputs.

During training, the Critic receives a separate noise-free privileged observation with the same history length:
\begin{equation}
\boldsymbol{o}_t^{\text{critic}} = [\boldsymbol{c}_t, \boldsymbol{c}_{t-1}, \boldsymbol{c}_{t-2}, \boldsymbol{c}_{t-3}, \boldsymbol{c}_{t-4}]
\end{equation}
where
\begin{equation}
\boldsymbol{c}_t = [\boldsymbol{v}_t^{\text{lin}}, \boldsymbol{\omega}_t, \boldsymbol{g}, \boldsymbol{v}_t^{\text{cmd}}, \boldsymbol{q}_t, \dot{\boldsymbol{q}}_t, \boldsymbol{a}_{t-1}, \Delta\boldsymbol{p}_t^{\text{zmp}}, d_t^{\text{ankle}}, \boldsymbol{\theta}_t^{\text{ankle}}].
\end{equation}
The additional privileged terms are the simulator ground-truth base linear velocity $\boldsymbol{v}_t^{\text{lin}} \in \mathbb{R}^3$, the centroidal ZMP offset with respect to the whole-body CoM in the world horizontal plane $\Delta\boldsymbol{p}_t^{\text{zmp}} \in \mathbb{R}^2$, the distance between the two ankle joints $d_t^{\text{ankle}} \in \mathbb{R}$, and the orientation angles of the two ankle joints with respect to the base's forward direction $\boldsymbol{\theta}_t^{\text{ankle}} \in \mathbb{R}^2$. These privileged terms provide the Critic with direct information about velocity tracking, support stability, swizzling geometry and foot heading during training.

\subsubsection{Action Space}

The policy outputs a dimensionless action vector $\boldsymbol{a}_t \in \mathbb{R}^{n_j}$ for all $25$ actuated DoFs.
At each control step, the action is scaled about the nominal configuration $\boldsymbol{q}_{\mathrm{default}}$ to form the desired joint positions
\begin{equation}
\boldsymbol{q}^d_t = \boldsymbol{q}_{\mathrm{default}} + \beta \boldsymbol{a}_t,
\end{equation}
where the action scale is set to $\beta=0.25$~rad.
A joint-level PD controller converts these targets into torques:
\begin{equation}
\boldsymbol{\tau}_t = \boldsymbol{K}_p(\boldsymbol{q}^d_t - \boldsymbol{q}_t) - \boldsymbol{K}_d\dot{\boldsymbol{q}}_t,
\end{equation}
where $\boldsymbol{K}_p$ and $\boldsymbol{K}_d$ are diagonal matrices of joint proportional and derivative gains, respectively.
Note that the eight passive wheels are excluded from the action vector; They receive neither position targets nor direct torque commands, and rotate only through ground contact dynamics.
Although the action space includes the arms, waist and neck to retain whole-body balance authority, only penalties are applied to softly regularize these joints around $\boldsymbol{q}_{\mathrm{default}}$ and suppress unnecessary upper-body motion.

\subsubsection{Reward Function Design}
\label{sec:reward}

Unlike walking rewards organized around alternating stance and swing contacts, the skating reward must maintain continuous rolling support, bounded lateral foot motion and low wheel axial slip. The instantaneous reward is $r_t=\sum_k w_k r_{k,t}$, where $r_{k,t}$ and $w_k$ denote the $k$-th reward component and its weight, respectively.
No kinematic reference trajectory, gait timing or state machine is imposed. Instead, swizzling is induced by four physical biases: inter-foot distance bounds, wheel axial-slip suppression, ZMP support stability and bilateral leg symmetry.
These biases are chosen to capture the physical mechanism of swizzling: inter-foot distance bounds define the opening-and-closing envelope, bilateral symmetry regularizes periodic leg coordination, wheel axial-slip suppression preserves the nonholonomic rolling behavior for propulsion, and ZMP terms maintain stable continuous support.

Let $d_{\mathrm{feet}}$ be the horizontal distance between the two feet. The distance-bound terms in Table~\ref{tab:rewards} constrain the swizzling envelope without prescribing a time-varying foot path. For the $i$-th passive wheel, $\mathbf{v}_i$, $\mathbf{a}_i$ and $F_i$ denote the wheel-center velocity, axle direction and normal contact force; the axial-slip term in Table~\ref{tab:rewards} is activated only for contacting wheels.

Support stability is enforced by ZMP terms. With whole-body CoM position $\mathbf{c}$, CoM acceleration $\mathbf{a}_c$, centroidal angular-momentum rate $\dot{\mathbf{h}}$ and net contact force $\mathbf{F}=M(\mathbf{a}_c-\mathbf{g})=[F_x,F_y,F_z]^\top$, the horizontal ZMP is
\begin{equation}
\mathbf{p}_{\mathrm{zmp}}^{xy}
=
\mathbf{c}^{xy}
+
\frac{1}{F_z}
\begin{bmatrix}
-c_zF_x-\dot{h}_y\\
-c_zF_y+\dot{h}_x
\end{bmatrix}.
\label{eq:zmp_estimation}
\end{equation}
When $F_z \le 0.1M\|\mathbf{g}\|$, the ZMP offset is set to zero. The support polygon $\mathcal{P}$ is built from contacting wheels, where the contacting wheels on the left and right feet are counted as $n_L$ and $n_R$. With signed ZMP margin $m$, safety margin $m_{\mathrm{safe}}=0.02$~m, smoothing scale $\sigma_m=0.03$~m and sigmoid $\sigma(\cdot)$, the support-valid reward is listed in Table~\ref{tab:rewards}, and the ZMP margin reward is
\begin{align}
\eta(m)
&= \sigma\left(\frac{m-m_{\mathrm{safe}}}{\sigma_m}\right) \notag\\
&\quad -0.3\tanh\left(\max\left(0,\frac{m_{\mathrm{safe}}-m}{\sigma_m}\right)\right), \notag\\
r_{\mathrm{zmp}}^{\mathrm{margin}}
&=
\begin{cases}
\eta(m), & |\mathcal{P}| \ge 3,\\
-0.3, & \mathrm{otherwise},
\end{cases}
\label{eq:reward_zmp_margin}
\end{align}

Finally, bilateral symmetry acts as a soft gait prior. With $\Delta\mathbf{q}=\mathbf{q}_L-\boldsymbol{S}\mathbf{q}_R$ and $\Delta\dot{\mathbf{q}}=\dot{\mathbf{q}}_L-\boldsymbol{S}\dot{\mathbf{q}}_R$, where $\boldsymbol{S}$ mirrors right-leg coordinates to the left-leg convention. The corresponding reward is summarized in Table~\ref{tab:rewards}.
Upper-body deviation penalties keep the arms, waist and head near a nominal posture while preserving balance authority. 

The remaining task, contact, posture, smoothness and energy terms follow standard locomotion regularization practice. All $22$ terms are summarized in Table~\ref{tab:rewards}.

\begin{table}[t]
\centering
\scriptsize
\setlength{\tabcolsep}{2pt}
\renewcommand{\arraystretch}{1.2}
\caption{Reward Function Terms}
\label{tab:rewards}
\begin{tabular}{@{}lcc@{}}
\toprule
\textbf{Reward Term} & \textbf{Formula} & \textbf{Weight} \\
\midrule
\multicolumn{3}{@{}l}{\textit{\textbf{Task Rewards}}} \\
Alive
& $\mathbf{1}_{\mathrm{alive}}$
& $0.15$ \\
Linear Velocity Track
& $\exp\left(-\frac{\|\mathbf{v}_{xy}^{\mathrm{cmd}}-\mathbf{v}_{xy}^{\mathrm{yaw}}\|^2}{0.25}\right)$
& $1.8$ \\
Angular Velocity Track
& $\exp\left(-\frac{(\omega_z^{\mathrm{cmd}}-\omega_z^{\mathrm{base}})^2}{0.25}\right)$
& $1.2$ \\
\addlinespace[3pt]
\multicolumn{3}{@{}l}{\textit{\textbf{Joint Penalties}}} \\
Arms Deviation
& $\sum_{\mathrm{arms}} |q_i-q_i^0|$
& $-0.4$ \\
Waist Deviation
& $\sum_{\mathrm{waist}} |q_i-q_i^0|$
& $-0.4$ \\
Head Deviation
& $|q_{\mathrm{head}}-q_{\mathrm{head}}^0|$
& $-1.0$ \\
Ankle Roll Deviation
& $\sum_{\mathrm{ankle\ roll}} |q_i-q_i^0|$
& $-0.2$ \\
\addlinespace[3pt]
\multicolumn{3}{@{}l}{\textit{\textbf{Posture Penalties}}} \\
Flat Orientation
& $g_{\mathrm{proj},x}^2+g_{\mathrm{proj},y}^2$
& $-2.0$ \\
Base Height
& $(h_{\mathrm{base}}-0.66)^2$
& $-2.0$ \\
\addlinespace[3pt]
\multicolumn{3}{@{}l}{\textit{\textbf{Swizzling Constraints}}} \\
Inter-foot Distance Lower Bound
& $\max(0,0.2-d_{\mathrm{feet}})$
& $-2.0$ \\
Inter-foot Distance Upper Bound
& $\max(0,d_{\mathrm{feet}}-0.5)$
& $-5.0$ \\
\addlinespace[3pt]
\multicolumn{3}{@{}l}{\textit{\textbf{Wheel and ZMP Rewards}}} \\
Wheel Axial Slip
& $\sum_i(\mathbf{v}_i\!\cdot\!\mathbf{a}_i)^2\mathbf{1}_{F_i>1\mathrm{N}}$
& $-0.1$ \\
ZMP Support Valid
& $\begin{cases}
1, & n_L \ge 2,\ n_R \ge 2\\
-1, & \mathrm{otherwise}
\end{cases}$
& $0.2$ \\
ZMP Support Margin
& Eq.~\eqref{eq:reward_zmp_margin}
& $0.5$ \\
\addlinespace[3pt]
\multicolumn{3}{@{}l}{\textit{\textbf{Symmetry Rewards}}} \\
Limb Symmetry
& $\begin{aligned}[t]
\frac{1}{1.5}\!\big[&
1.0\exp\!\left(-\frac{\|\Delta\mathbf{q}\|^2}{0.25}\right)\\[-0.5em]
&+0.5\exp\!\left(-\frac{\|\Delta\dot{\mathbf{q}}\|^2}{4.0}\right)
\big]
\end{aligned}$
& $0.5$ \\
\addlinespace[3pt]
\multicolumn{3}{@{}l}{\textit{\textbf{Contact Penalties}}} \\
Undesired Contacts
& $\sum_{\substack{\mathrm{body}\\\notin \mathrm{wheel}}} \mathbf{1}_{F_{\mathrm{contact}}>1\mathrm{N}}$
& $-1.0$ \\
\addlinespace[3pt]
\multicolumn{3}{@{}l}{\textit{\textbf{Regularization Penalties}}} \\
Angular Velocity XY
& $\omega_x^2+\omega_y^2$
& $-0.05$ \\
Joint Velocity
& $\|\dot{\mathbf{q}}\|^2$
& $-0.001$ \\
Joint Acceleration
& $\|\ddot{\mathbf{q}}\|^2$
& $-2.5\times10^{-7}$ \\
Action Rate
& $\|\mathbf{a}_t-\mathbf{a}_{t-1}\|^2$
& $-0.05$ \\
Energy
& $\sum_i |\tau_i| |\dot{q}_i|$
& $-2\times10^{-5}$ \\
Joint Position Limits
& $\begin{aligned}[t]
&\sum_i [\max(0,q_{\min,i}-q_i)\\[-0.5em]
&\quad+\max(0,q_i-q_{\max,i})]
\end{aligned}$
& $-5.0$ \\
\bottomrule
\end{tabular}
\end{table}

\subsubsection{Sim-to-Real Transfer}

The sim-to-real transfer gap constitutes one of the key challenges in reinforcement learning for physical robots. To improve transfer robustness, domain randomization is applied during training by perturbing wheel-ground contact properties, robot inertial parameters, actuator gains and external disturbances. The randomized quantities and their ranges are listed in Table~\ref{tab:domain_randomization}.

\begin{table}[ht]
\centering
\caption{Domain Randomization Parameters}
\label{tab:domain_randomization}
\begin{tabular}{@{}p{0.48\columnwidth}p{0.43\columnwidth}@{}}
\toprule
\textbf{Term} & \textbf{Value} \\
\midrule
\multicolumn{2}{@{}l}{\textit{\textbf{Material Physics - Wheels}}} \\
Static Friction Coefficient & $\mathcal{U}(0.1, 0.8)$ \\
Dynamic Friction Coefficient & $\mathcal{U}(0.1, 0.4)$ \\
\addlinespace
\textit{\textbf{Base Mass Offset}} & $\mathcal{U}(-1.0, 3.0)$ kg added to pelvis \\
\addlinespace
\multicolumn{2}{@{}l}{\textit{\textbf{CoM Offset - Pelvis and Torso}}} \\
$x/y/z$ axes & $\mathcal{U}(-0.01, 0.01)$ m \\
\addlinespace
\multicolumn{2}{@{}l}{\textit{\textbf{Actuated Joint Parameters - Non-wheel Joints}}} \\
Proportional Gain Scale & $\mathcal{U}(0.9, 1.1)$ \\
Derivative Gain Scale & $\mathcal{U}(0.9, 1.1)$ \\
Motor Zero Offset & $\mathcal{U}(-0.02, 0.02)$ rad \\
\addlinespace
\multicolumn{2}{@{}l}{\textit{\textbf{External Disturbance}}} \\
Push Velocity $x$ & $\mathcal{U}(-0.4, 0.5)$ m/s \\
Push Velocity $y$ & $\mathcal{U}(-0.8, 0.8)$ m/s \\
Push Interval & $5.0$ s \\
\bottomrule
\end{tabular}
\end{table}

\section{Experiments and Results}
\label{Experiments}
To evaluate the performance of inline skate swizzles, nominal bipedal walking gaits are also tested on SKATER with the wheel mechanisms replaced by normal rigid flat feet. The walking baseline is trained using the same RL training pipeline as the skating policy, including the same simulator, PPO implementation, policy network structure, action interface and PD control layer. The reward functions are task-specific: the walking policy uses standard bipedal locomotion rewards for velocity tracking, posture stabilization, regular foot contacts and smooth joint motion, whereas the skating policy additionally incorporates rolling-contact and nonholonomic constraints introduced in Section~\ref{sec:reward}.
For both swizzling and walking gaits, the learned policy is first deployed in the MuJoCo physics engine~\cite{todorov2012mujoco} for sim-to-sim tests given its high-precision contact force simulation, and then deployed on SKATER for sim-to-real evaluation.

\subsection{Simulation Results}

\subsubsection{Impact Force Analysis}

\begin{figure}[t]
    \centering
    \includegraphics[width=1.0\columnwidth]{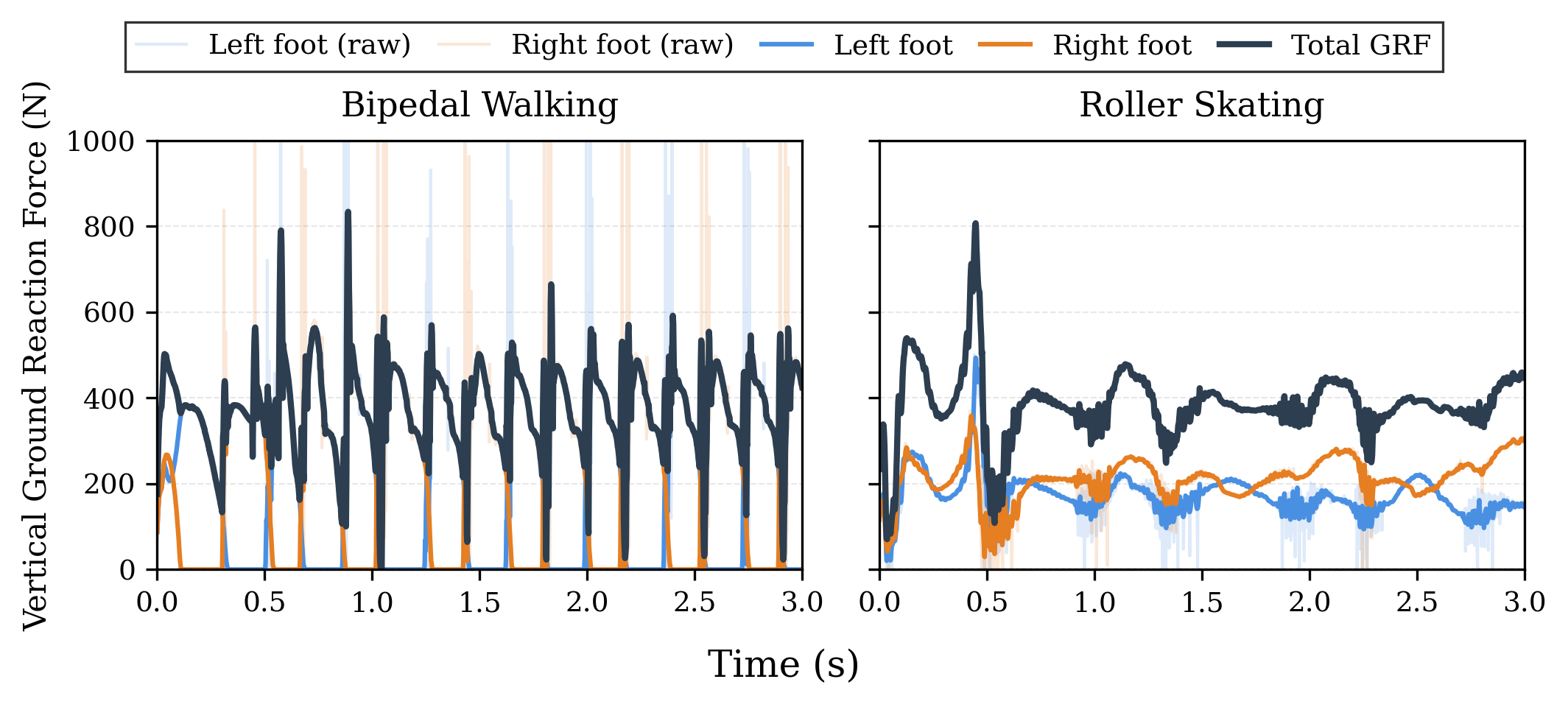}
    \caption{Comparison of vertical GRF profiles.}
    \label{fig:foot_force_comparison}
\end{figure}

Fig.~\ref{fig:foot_force_comparison} presents a temporal comparison of vertical Ground Reaction Forces (GRF) between the swizzling and the walking gait during the first $3$~s at a commanded speed of $1$ m/s. Given the weight of SKATER being approximately $372.4$~N, it can be seen that nominal walking generates pronounced impact peaks at each heel strike, reaching $1.5$-$2.0$ times body weight during the approximately $50$-$100$~ms time window, imposing substantial loads on the joints. In contrast, the force profile of the swizzling gait exhibits smooth periodic variations. Excluding the startup portion where the robot must exert force to accelerate, the peak forces are reduced to $1.1$-$1.3$ times body weight with significantly decreased force variation rates.

To further quantify the impact severity difference, we define an Impact Intensity metric as:
\begin{equation}
I = \frac{1}{T}\int_{t_0}^{t_f} \left|\frac{\mathrm{d}F_z}{\mathrm{d}t}\right| \mathrm{d}t
\end{equation}
As a result, the Impact Intensity for walking is $I_\text{bipedal}=2283.17$ N/s, while that for swizzling is $I_{\text{roller}}=1135.93$ N/s, which is an reduction of $50.25\%$ and indicates the advantage of skating locomotion in reducing joint impacts. Additional tests at speed commands of $0.5$, $1.5$ and $2.0$ m/s observe consistent reductions of $44.97\%$, $50.18\%$ and $44.64\%$, respectively. 

\subsubsection{Velocity Tracking Performance}

\begin{figure}[t]
    \centering
    \includegraphics[width=1.0\columnwidth]{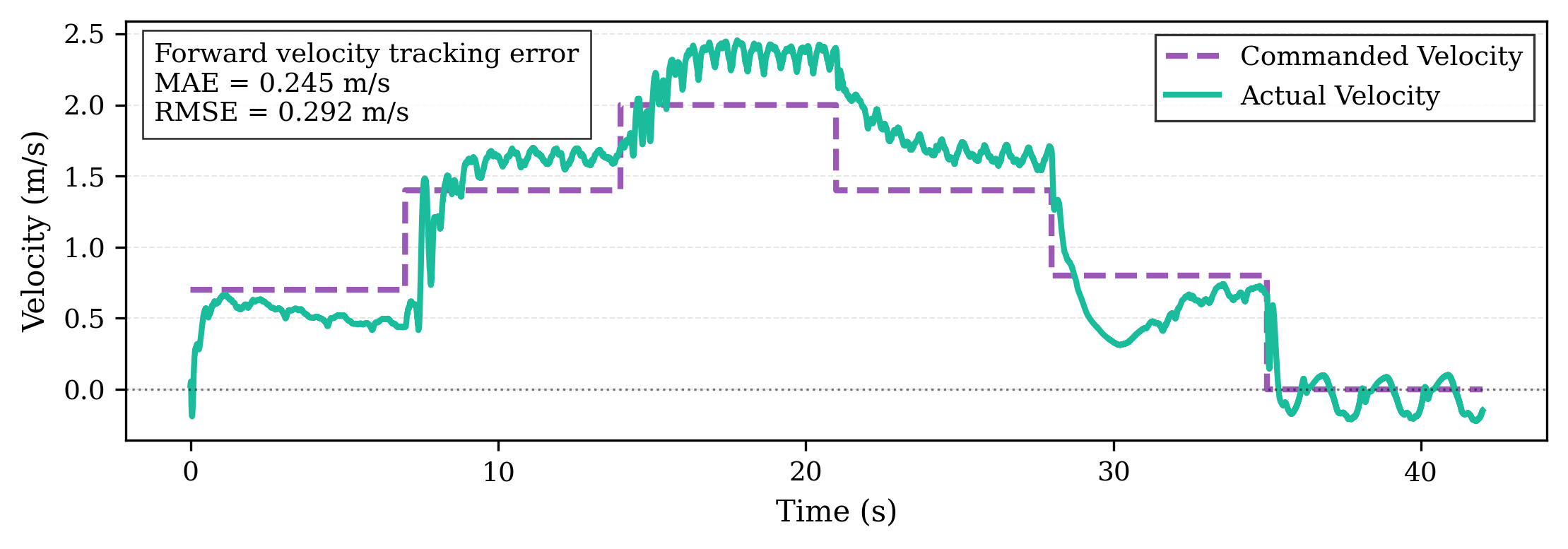}
    \caption{Velocity tracking performance of SKATER following step commands. }
    \label{fig:velocity_comparison}
\end{figure}

To evaluate SKATER's velocity tracking performance, the desired forward velocity sequence was set to $0.7$, $1.4$, $2.0$, $1.4$, $0.7$, $0.0$~m/s in MuJoCo, with a switching interval of $7.0$~s. 
The results presented in Fig.~\ref{fig:velocity_comparison} show reasonable velocity tracking, with accelerating transitions producing short speed spikes and decelerating exhibiting a slower response. These behaviors reflect the limited braking and speed-regulation authority of passive-wheel rolling contact compared with conventional foot-ground stance control.

\subsubsection{Gait Pattern Visualization}

\begin{figure}[t]
    \centering
    \includegraphics[width=0.8\columnwidth]{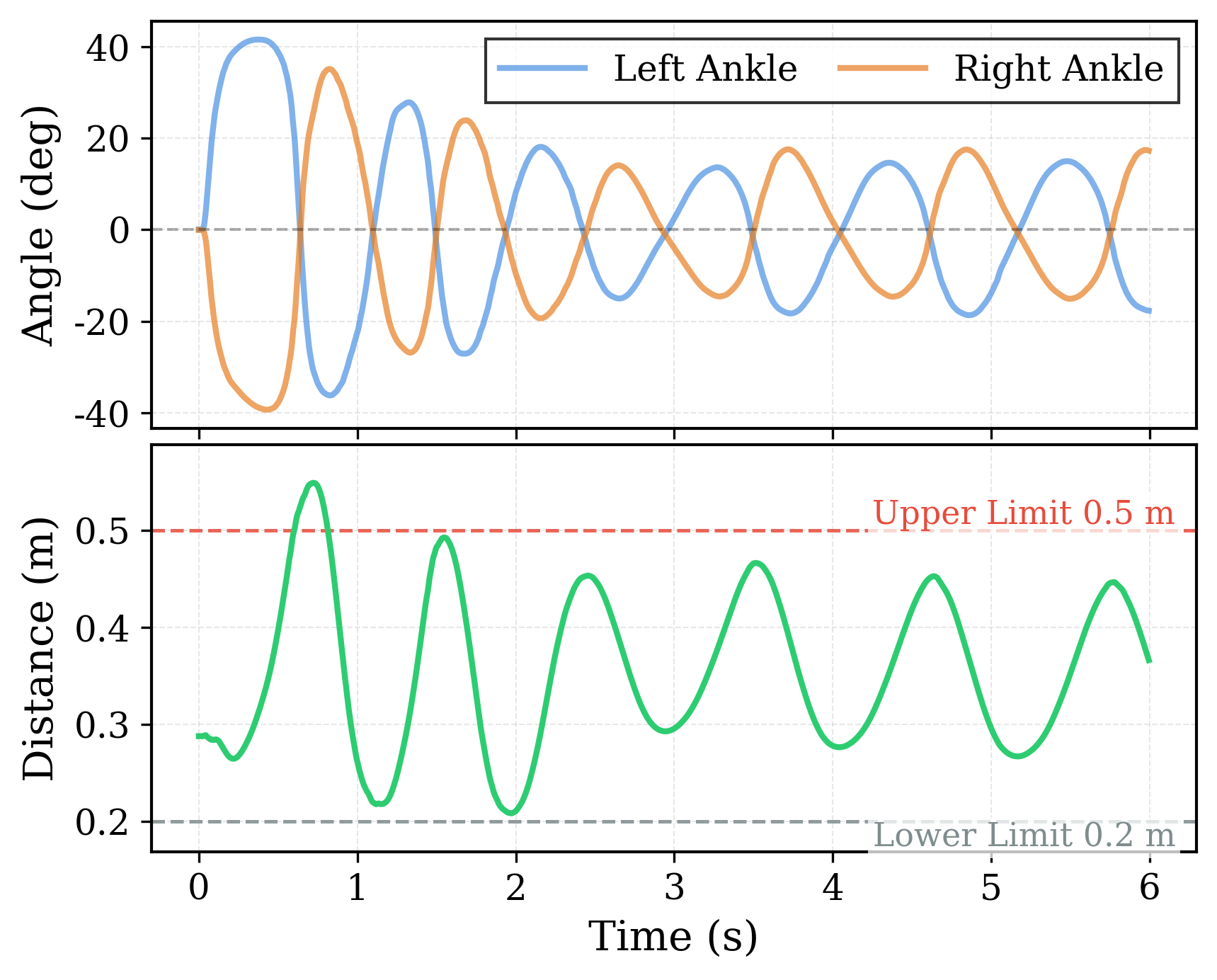}
    \caption{Temporal analysis of the learned swizzling gait: ankle joint angle profiles and inter-foot distance variation with the reward bounds indicated.}
    \label{fig:foot_trajectories}
\end{figure}

For a swizzling gait, Fig.~\ref{fig:foot_trajectories} shows the temporal coordination between the ankle motions and the opening-and-closing foot distance. The ankle angles exhibit periodic left-right coordination, while the inter-foot distance follows the desired swizzling envelope after a short initial transient.
These results demonstrate that through appropriately designed physical constraints and reward terms, DRL can form and optimize swizzling gaits without reference motion data or an explicit state machine. This data-driven approach provides a practical way for inline skating locomotion control of humanoid robots with passive wheels.

\subsection{Ablation Study}
\label{Ablation Study}

To identify the contribution of individual reward terms, we conduct ablations over five random seeds. All policies are evaluated in Isaac Lab under the same nominal protocol on a fixed $18$-command grid with $v_x \in \{-1.0,-0.5,0.5,1.0,1.5,2.0\}$~m/s, $v_y=0$, and $\omega_z \in \{-0.5,0,0.5\}$~rad/s. For each command, we collect $100$ completed episodes using $32$ parallel environments, with physical-parameter domain randomization disabled. The results are presented in Table~\ref{tab:reward_ablation}. The task survival rate represents the fraction of episodes completed without falling, while worst-command survival is the minimum across command pairs. Because survival alone does not establish the desired swizzling pattern, we also report command-tracking RMSE and joint mechanical energy per meter. Diagnostic metrics follow the skating-specific rewards: ZMP inside rate and margin measure support inclusion and clearance, inter-foot bound violations and leg symmetry quantify gait geometry, wheel axial slip measures nonholonomic slip, and wheel contact validity checks whether ZMP support has sufficient contacts.

\begin{table*}[t]
\centering
\caption{Reward term ablations. Values are reported as mean $\pm$ standard deviation over five random seeds.}
\label{tab:reward_ablation}
\scriptsize
\setlength{\tabcolsep}{3pt}
\renewcommand{\arraystretch}{0.92}
\resizebox{\textwidth}{!}{
\begin{tabular}{lcccccc}
\toprule
\multicolumn{7}{@{}l}{\textbf{(a) Main performance metrics}} \\
\midrule
Method
& Task Survival (\%) $\uparrow$
& Worst-cmd. Survival (\%) $\uparrow$
& Fall Rate (\%) $\downarrow$
& Forward Vel. RMSE (m/s) $\downarrow$
& Yaw-rate RMSE (rad/s) $\downarrow$
& Energy per Meter $\downarrow$ \\
\midrule
Full reward
& \underline{94.62 $\pm$ 0.75}
& \underline{64.80 $\pm$ 9.09}
& \underline{5.38 $\pm$ 0.75}
& 0.236 $\pm$ 0.001
& \textbf{0.194 $\pm$ 0.001}
& 113.95 $\pm$ 0.27 \\
w/o symmetry reward
& \textbf{99.17 $\pm$ 0.41}
& \textbf{93.00 $\pm$ 3.08}
& \textbf{0.83 $\pm$ 0.41}
& 0.257 $\pm$ 0.000
& 0.329 $\pm$ 0.000
& 140.13 $\pm$ 0.11 \\
w/o ZMP reward
& 87.22 $\pm$ 1.13
& 26.40 $\pm$ 2.70
& 12.78 $\pm$ 1.13
& \textbf{0.196 $\pm$ 0.001}
& 0.255 $\pm$ 0.002
& \textbf{81.91 $\pm$ 0.61} \\
w/o ZMP margin reward
& 90.00 $\pm$ 0.93
& 0.00 $\pm$ 0.00
& 10.00 $\pm$ 0.93
& 0.241 $\pm$ 0.003
& \underline{0.251 $\pm$ 0.000}
& \underline{90.92 $\pm$ 0.84} \\
w/o foot-distance reward
& 86.19 $\pm$ 0.93
& 0.00 $\pm$ 0.00
& 13.81 $\pm$ 0.93
& 0.273 $\pm$ 0.001
& 0.326 $\pm$ 0.004
& 154.34 $\pm$ 1.15 \\
w/o wheel axial-slip reward
& 91.40 $\pm$ 0.30
& 33.20 $\pm$ 9.18
& 8.60 $\pm$ 0.30
& \underline{0.235 $\pm$ 0.004}
& 0.285 $\pm$ 0.001
& 167.47 $\pm$ 0.15 \\
\bottomrule
\end{tabular}
}
\vspace{0.2em}
\resizebox{\textwidth}{!}{
\begin{tabular}{lcccccc}
\toprule
\multicolumn{7}{@{}l}{\textbf{(b) Diagnostic metrics}} \\
\midrule
Method
& ZMP Inside Rate (\%) $\uparrow$
& ZMP Margin Reward $\uparrow$
& Inter-foot Bound Viol. (\%) $\downarrow$
& Leg Symmetry Score $\uparrow$
& Wheel Axial Slip $\downarrow$
& Wheel Contact Validity (\%) $\uparrow$ \\
\midrule
Full reward
& \textbf{99.47 $\pm$ 0.02}
& \textbf{0.768 $\pm$ 0.001}
& 4.27 $\pm$ 0.07
& \textbf{0.894 $\pm$ 0.000}
& 0.191 $\pm$ 0.000
& \textbf{97.11 $\pm$ 0.08} \\
w/o symmetry reward
& \underline{99.28 $\pm$ 0.01}
& \underline{0.740 $\pm$ 0.000}
& \textbf{2.09 $\pm$ 0.02}
& 0.233 $\pm$ 0.000
& 0.226 $\pm$ 0.000
& \underline{95.33 $\pm$ 0.04} \\
w/o ZMP reward
& 62.23 $\pm$ 0.09
& 0.000 $\pm$ 0.002
& 8.01 $\pm$ 0.07
& 0.756 $\pm$ 0.001
& \textbf{0.063 $\pm$ 0.000}
& 39.68 $\pm$ 0.13 \\
w/o ZMP margin reward
& 97.44 $\pm$ 0.10
& 0.640 $\pm$ 0.002
& \underline{3.54 $\pm$ 0.09}
& \underline{0.877 $\pm$ 0.000}
& 0.130 $\pm$ 0.000
& 92.70 $\pm$ 0.11 \\
w/o foot-distance reward
& 95.36 $\pm$ 0.03
& 0.643 $\pm$ 0.001
& 76.14 $\pm$ 0.41
& 0.843 $\pm$ 0.001
& \underline{0.095 $\pm$ 0.000}
& 87.05 $\pm$ 0.14 \\
w/o wheel axial-slip reward
& 96.63 $\pm$ 0.13
& 0.654 $\pm$ 0.002
& 56.49 $\pm$ 0.28
& 0.651 $\pm$ 0.000
& 2.541 $\pm$ 0.022
& 83.41 $\pm$ 0.20 \\
\bottomrule
\end{tabular}
}
\end{table*}

The ablation results should be interpreted as a multi-objective swizzling-quality analysis rather than a single survival leaderboard. The best result in each column of Table~\ref{tab:reward_ablation} is shown in bold, while the second-best result is underlined. The full reward does not maximize survival alone; instead, it achieves the best yaw-rate tracking, ZMP inside rate, ZMP margin reward, leg symmetry score and wheel contact validity, indicating a balanced trade-off among tracking, support stability, gait regularity and rolling contact. Removing the ZMP reward degrades the ZMP inside rate from $99.47\%$ to $62.23\%$ and reduces the worst-command survival rate from $64.80\%$ to $26.40\%$, showing its role in stable support. Removing the ZMP margin reward leads to zero worst-command survival, while removing the foot-distance reward increases inter-foot bound violation to $76.14\%$. Removing the wheel axial-slip reward increases axial slip from $0.191$ to $2.541$ and yields the highest energy consumption, confirming its role in enforcing nonholonomic rolling. The symmetry ablation is particularly revealing that, although removing symmetry gives the highest nominal survival rate, it substantially degrades yaw-rate tracking, energy efficiency and leg symmetry. This indicates that the symmetry term is not mainly a fall-prevention term, but a gait-quality term that prevents the policy from surviving through asymmetric or shuffling motions that no longer match the intended swizzling primitive.

\begin{figure}[t]
    \centering
    \includegraphics[width=1.0\columnwidth]{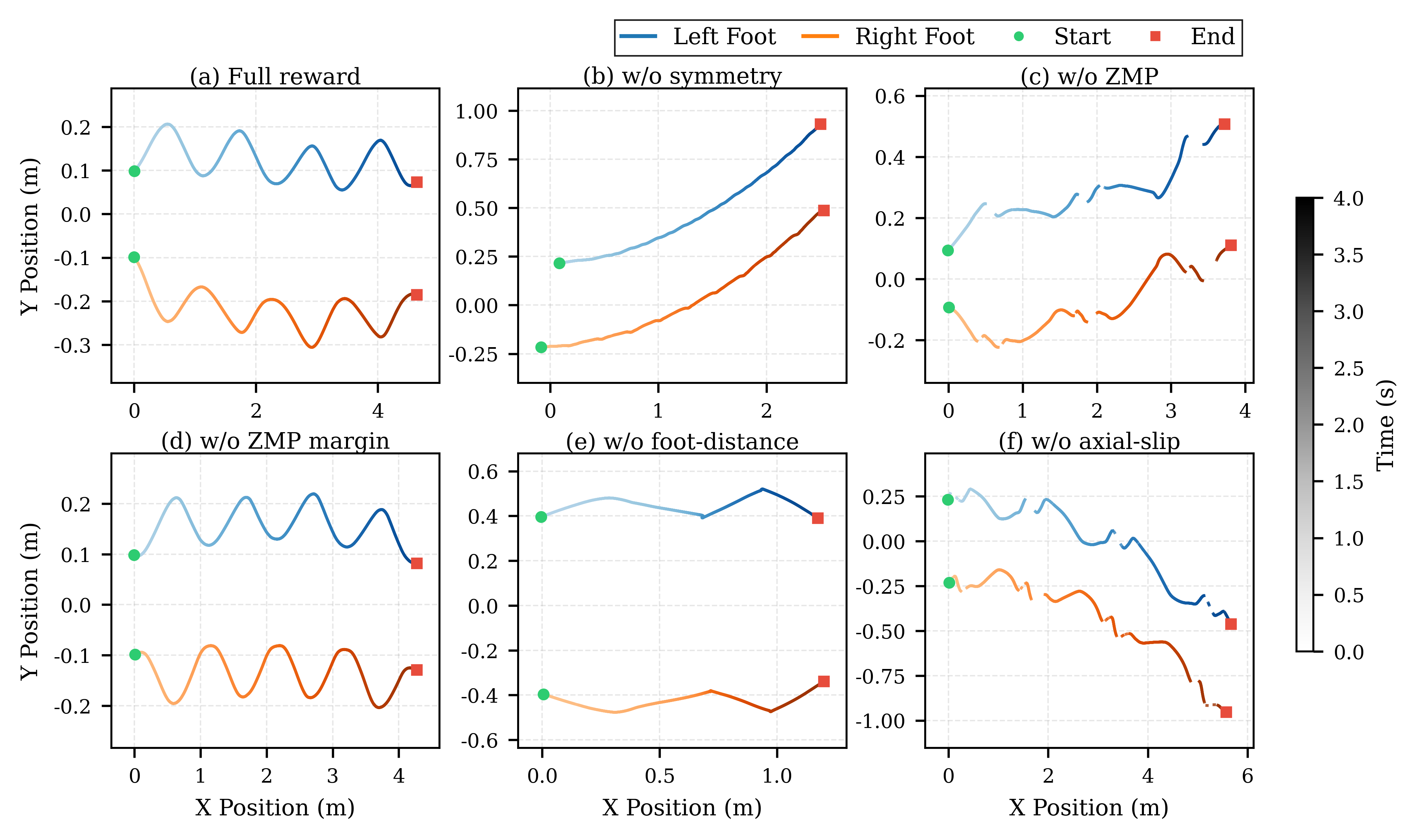}
    \caption{Horizontal foot trajectories under representative reward term ablations and their motion-level failure modes.}
    \label{fig:reward_ablation_foot_trajectories}
\end{figure}

Fig.~\ref{fig:reward_ablation_foot_trajectories} links the metrics to motion-level failures through horizontal foot trajectories. The full reward and the variant without the ZMP margin reward retain swizzling-like trajectories; removing symmetry produces low-clearance shuffling, removing the ZMP or axial-slip reward induces stepping-like lift-off, and removing the foot-distance reward breaks the desired swizzling geometry.


\subsection{Experimental Results}

The learned control policy is then transferred to the physical robot for real world evaluation. The policy is run at $50$~Hz through the onboard computer to show its efficacy.

\subsubsection{Real-World Mobility Evaluation}

\begin{figure}[t]
    \centering
    \subfloat[]{
        \includegraphics[width=0.96\columnwidth]{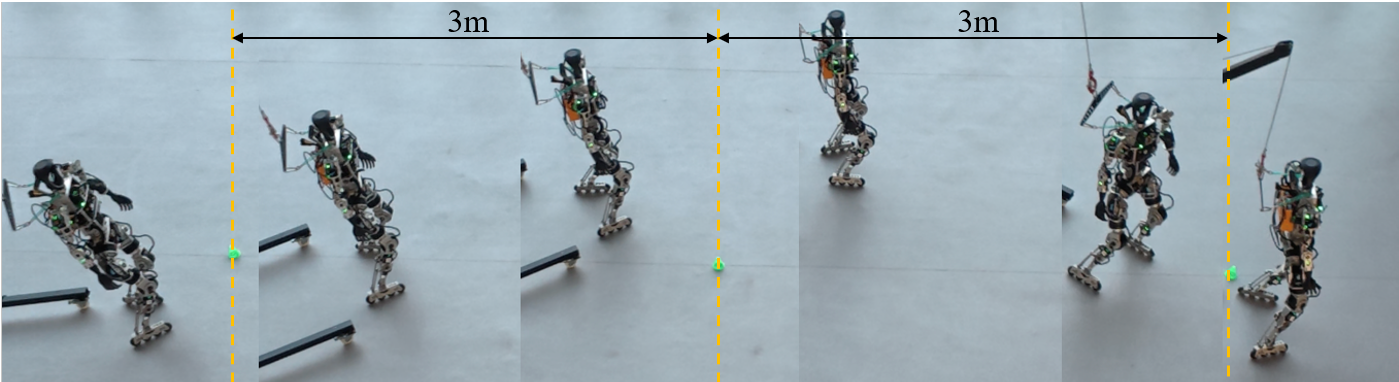}
        \label{fig:turning}
    }\\
    \subfloat[]{
        \includegraphics[width=0.96\columnwidth]{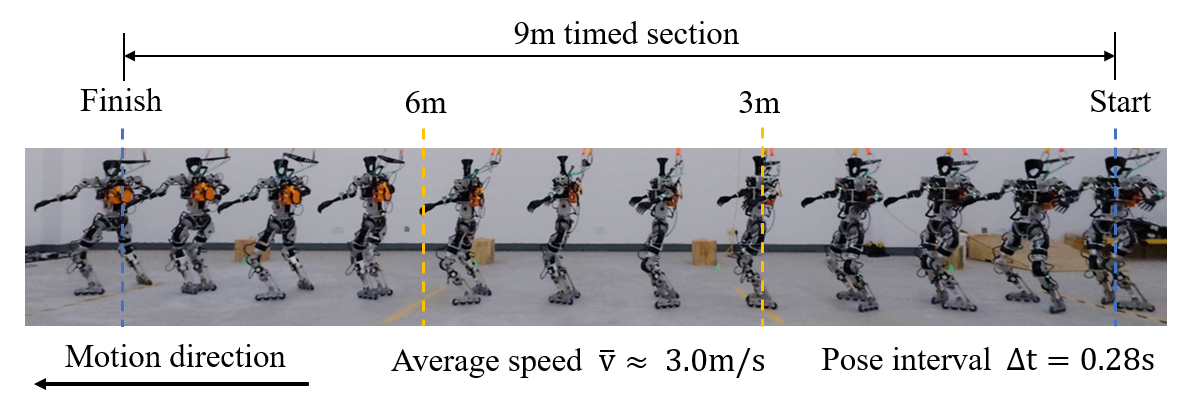}
        \label{fig:high_speed_test}
    }
    \caption{Real-world mobility validation: (a) obstacle-crossing swizzling and (b) high-speed flat-ground traversal.}
    \label{fig:real_world_mobility}
\end{figure}

The real-world mobility of SKATER is evaluated from three aspects: basic omnidirectional locomotion, high-speed flat-ground traversal and multi-surface robustness. Forward, backward and obstacle-crossing swizzling trials were conducted from rest, followed by stable braking through a pigeon-toed stance. As shown in Fig.~\ref{fig:real_world_mobility}(a), the obstacle-crossing trial validates turning maneuverability. For high-speed traversal, five repeated trials were performed over a marked $9$~m straight section, with traversal time recorded by stopwatch and cross-checked using video timestamps. The trials reported a traversal time of $3.01\pm0.01$~s and an average traversal speed of $2.99\pm0.01$~m/s over the measured section, with Fig.~\ref{fig:real_world_mobility}(b) showing a representative trial.
Additionally, to evaluate the robot's adaptability to surfaces with distinct contact and rolling-resistance characteristics, tests were conducted on tile, rubber and gravel pavement. On each surface, the robot performed $10$ repeated $10$-m swizzling trials and achieved $10/10$ success. A trial was considered successful if the robot completed the course without falling or human intervention.

\subsubsection{Joint Impact Load Evaluation}

To quantitatively assess the mechanical impact of walking and swizzling gaits on robot joints, joint torque data were collected from SKATER under both gaits at a commanded speed of $1$~m/s. Peak torque is employed as the evaluation metric, focusing mainly on the dynamic characteristics of the $12$ lower-limb joints. 
Table~\ref{tab:joint_operation_comparison} presents the detailed results, indicating that the swizzling gait reduces peak torques across most lower-limb joints. The most pronounced reductions are observed in the sagittal-plane joints: the left and right hip pitch torques are reduced by $69.53\%$ and $73.45\%$, respectively;
the left and right ankle pitch torques are reduced by $70.71\%$ and $61.66\%$, respectively.
This is attributed to the continuous sliding motion in the swizzling gait, which avoids the periodic impacts at touchdown during walking. These results indicate that swizzling can reduce peak impact loads on key lower-limb joints.

\begin{table}[t]
    \caption{Comparison of peak joint torques between swizzling and walking}
    \label{tab:joint_operation_comparison}
    \centering
    \begin{tabular}{l c c c}
        \hline
        \textbf{Joint Name} & \textbf{Swizzling(N·m)} & \textbf{Walking(N·m)} & \textbf{Difference (\%)} \\
        \hline
        L Hip Pitch    & $20.86\pm6.22$ & $68.47\pm1.91$ & $-$69.53 \\
        L Hip Roll     & $56.22\pm4.08$ & $60.62\pm2.25$ & $-$7.26  \\
        L Hip Yaw      & $23.18\pm1.39$ & $33.82\pm1.19$ & $-$31.46 \\
        L Knee         & $49.48\pm2.98$ & $51.76\pm6.84$ & $-$4.41  \\
        L Ankle Pitch  & $13.58\pm2.51$ & $46.37\pm6.10$ & $-$70.71 \\
        L Ankle Roll   & $11.80\pm2.77$ & $11.08\pm1.84$ & $+$6.55  \\
        \hline
        R Hip Pitch    & $17.87\pm3.31$ & $67.32\pm0.92$ & $-$73.45 \\
        R Hip Roll     & $59.85\pm7.91$ & $85.04\pm9.45$ & $-$29.63 \\
        R Hip Yaw      & $23.50\pm3.19$ & $40.23\pm3.62$ & $-$41.58 \\
        R Knee         & $42.88\pm8.85$ & $45.37\pm1.99$ & $-$5.49  \\
        R Ankle Pitch  & $15.21\pm2.21$ & $39.67\pm0.85$ & $-$61.66 \\
        R Ankle Roll   & $11.78\pm1.36$ & $12.45\pm0.97$ & $-$5.37  \\
        \hline
    \end{tabular}
\end{table}

\subsubsection{Joint-Mechanical Cost of Transport Comparison}

For energy-efficiency evaluation, the walking baseline replaces the passive-wheel feet with rigid flat feet while preserving the same sensing and actuation stack. 
With each inline skate module being $0.24$~kg heavier than the rigid flat foot, the skating configuration is $0.48$~kg heavier than the walking configuration, corresponding to $1.26\%$ of its $38$~kg total mass. 
Thus, the reported energy reduction cannot be attributed to a lighter skating setup. Each locomotion mode performs $5$ repeated straight-line trials over $15$~m at a commanded speed of $1$~m/s. Since computing and sensing power are comparable, we report mechanical CoT~\cite{alqaham2024energy}:
\begin{equation}
\text{CoT} = \frac{\sum_{i=1}^{n}\left(\int_{t_0}^{t_f}|\tau_i \cdot \omega_i| \, \mathrm{d}t \right)}{M \cdot g \cdot D}
\end{equation}
where $\tau_i$ and $\omega_i$ are the torque and angular velocity of joint $i$, $M$ is the robot mass, $g$ is the gravitational acceleration, $D$ is the travel distance, and $n$ is the number of joints.

\begin{figure}[t]
    \centering
    \includegraphics[width=1.0\columnwidth]{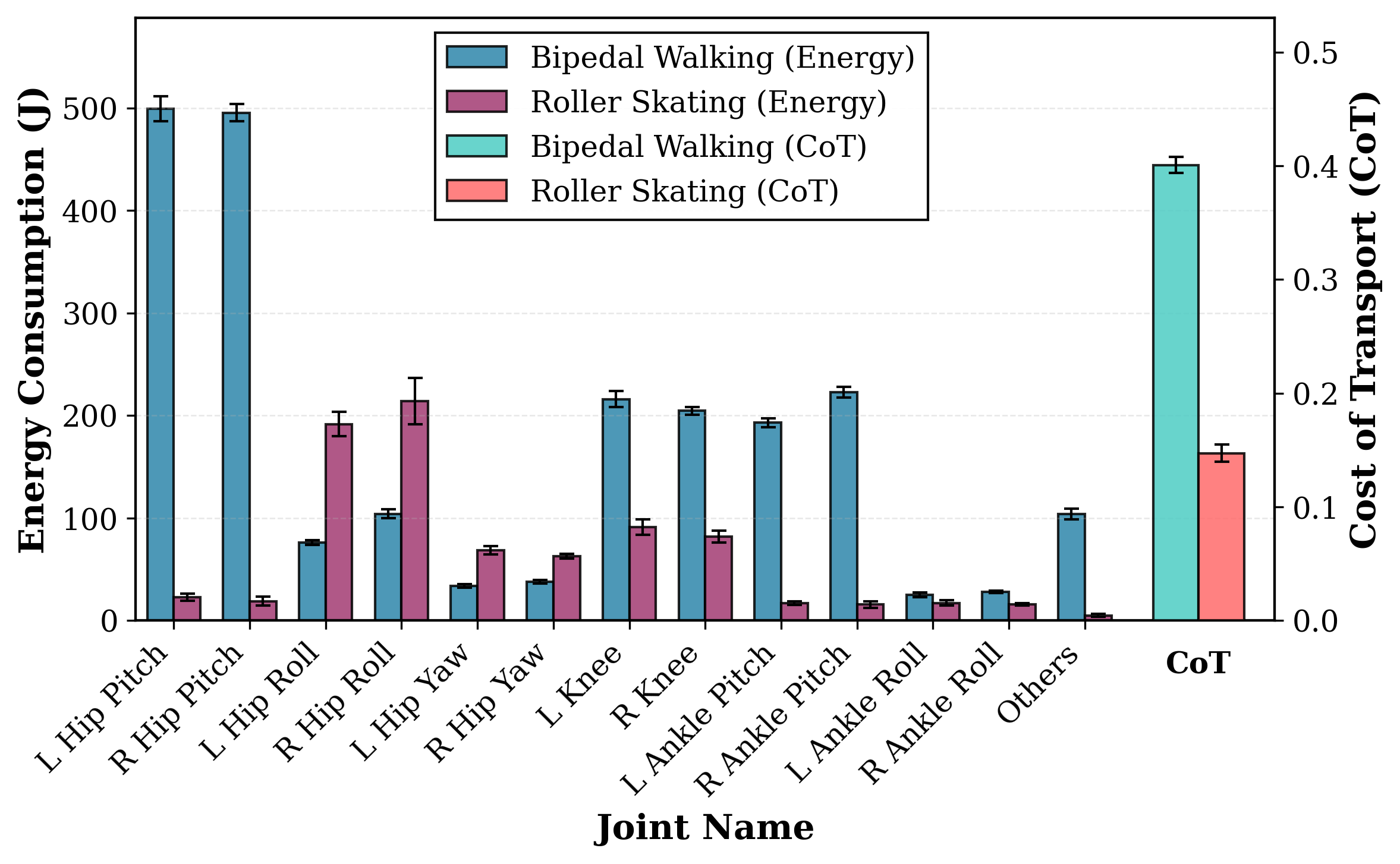}
    \caption{Energy efficiency results: joint-level energy consumption and mechanical Cost of Transport comparison between swizzling and walking gaits.}
    \label{fig:cot_comparsion}
\end{figure}

Fig.~\ref{fig:cot_comparsion} shows that swizzling reduces mechanical CoT by $63.34\%$ compared with walking. The main savings occur in sagittal-plane joints: energy consumption of hip pitch decreases by approximately $95.39\%$ and ankle pitch by $92.29\%$, while other joints show smaller changes. These results indicate that continuous rolling contact reduces the actuation demand associated with discrete stance-swing walking.

\section{Conclusions}
\label{Conclusion}

This work demonstrates that swizzling is an effective entry-point gait primitive for passive-wheel humanoid inline skating. Built on the $33$-DoF SKATER humanoid robot, the proposed reference-free deep reinforcement learning framework synthesizes swizzling without motion-capture demonstrations or manually designed reference motions. Rather than tracking a prescribed motion or using explicit gait-phase scheduling, the policy uses skating-specific physical constraints, including ZMP support, inter-foot geometry, bilateral symmetry and wheel axial-slip suppression, to guide whole-body motion learning under passive nonholonomic wheel-ground contact. Compared with a walking baseline on the same robot but with rigid flat feet, the learned swizzling gait reduces GRF-based Impact Intensity by $50.25\%$ in simulation and mechanical CoT by $63.34\%$ in real-world trials. Hardware experiments across different surfaces and high-speed traversal trials on flat ground further indicate the robustness and speed potential of the learned policy. These results suggest that continuous bilateral-support swizzling provides a practical foundation for low-impact and energy-efficient humanoid inline skating.

The current study remains limited to swizzling gaits with continuous contact of both feet. Lateral drift during straight-line skating and delayed velocity regulation during acceleration and braking remain observable, reflecting the limited steering and braking authority of passive inline wheels. Future work will extend this foundation toward more dynamic skating skills, such as push-step and single-support skating, integrate perception for autonomous navigation, and explore switchable wheel/flat-foot mechanisms to improve terrain adaptability.

\bibliographystyle{IEEEtran}
\bibliography{ref}

\end{document}